\documentclass[10pt,twocolumn,letterpaper]{article}

\usepackage{cvpr}              

\definecolor{cvprblue}{rgb}{0.21,0.49,0.74}
\usepackage[pagebackref,breaklinks,colorlinks,allcolors=cvprblue]{hyperref}

\title{BiCLIP: Domain Canonicalization via Structured Geometric Transformation}

\author{Pranav Mantini\\
University of Houston\\
Department of Computer Science\\
{\tt\small pmantini@uh.edu}
\and
Shishir K. Shah\\
The University of Oklahoma\\
School of Computer Science\\
{\tt\small sshah@ou.edu}
}

\begin{document}
\maketitle
\begin{abstract}
Recent advances in vision-language models (VLMs) have demonstrated remarkable zero-shot capabilities, yet adapting these models to specialized domains remains a significant challenge. Building on recent theoretical insights suggesting that independently trained VLMs are related by a canonical transformation, we extend this understanding to the concept of domains. We hypothesize that image features across disparate domains are related by a canonicalized geometric transformation that can be recovered using a small set of anchors. Few-shot classification provides a natural setting for this alignment, as the limited labeled samples serve as the anchors required to estimate this transformation. Motivated by this hypothesis, we introduce BiCLIP, a framework that applies a targeted transformation to multimodal features to enhance cross-modal alignment. Our approach is characterized by its extreme simplicity and low parameter footprint. Extensive evaluations across 11 standard benchmarks, including EuroSAT, DTD, and FGVCAircraft, demonstrate that BiCLIP consistently achieves state-of-the-art results. Furthermore, we provide empirical verification of existing geometric findings by analyzing the orthogonality and angular distribution of the learned transformations, confirming that structured alignment is the key to robust domain adaptation. Code is available at \url{https://github.com/QuantitativeImagingLaboratory/BilinearCLIP}

\end{abstract}    
\section{Introduction}
\label{sec:intro}
\textbf{Few-shot classification}: Image classification is a fundamental problem in computer vision, where the objective is to categorize images into predefined semantic classes. Few-shot classification is a more restrictive formulation where a limited number of samples are available for training—a constraint more representative of real-world scenarios. Recent methods focus on leveraging foundational models trained on large-scale generalized datasets and adapting their learned representations for domain-specific downstream tasks~\cite{manzoor2024multimodalityrepresentationlearningsurvey, gu2022openvocabularyobjectdetectionvision, zhou2022detectingtwentythousandclassesusing, maaz2022classagnosticobjectdetectionmultimodal}. Vision language models (VLMs)~\cite{yao2021filipfinegrainedinteractivelanguageimage, yuan2021florencenewfoundationmodel, zhai2022litzeroshottransferlockedimage}, specifically based on the Contrastive Language-Image Pre-training (CLIP, SigLIP) ~\cite{radford2021learningtransferablevisualmodels, zhai2023sigmoidlosslanguageimage}, have demonstrated that high-dimensional features extracted from such models are inherently discriminative \cite{radford2021learningtransferablevisualmodels}. Attributing to this, they exhibit remarkable zero-shot capabilities; however, their performance degrades in domain-specific fine-grained scenarios where the models struggle to distinguish between similar intra-class objects.

\textbf{Principles for Few-Shot Adaptation of VLMs}:  While contrastive VLMs like CLIP and SigLIP provide a robust starting point, the transition to few-shot classification requires an adaptation strategy that is computationally efficient and geometrically well defined. Recent methods~\cite{zhou2022conditionalpromptlearningvisionlanguage, Zhou_2022, khattak2023maplemultimodalpromptlearning, gondal2024domain} have focused on adapter-based strategies that refine the high-dimensional feature space without deconstructing the foundational knowledge acquired during large-scale pre-training. 
These models converge upon the following shared design principles: 
\begin{enumerate}
    \item \textbf{Minimal Parameter Overhead}: Majority of adaptation strategies refrain from initializing parameter-intensive architectures and favor techniques for Parameter-Efficient Fine-Tuning (PEFT)~\cite{xu2023parameterefficientfinetuningmethodspretrained}. Prompt Learning (CoOp, CoCoOp, MaPLe)~\cite{zhou2022conditionalpromptlearningvisionlanguage, Zhou_2022, khattak2023maplemultimodalpromptlearning}, a largely successful approach, only learns a few tokens ($<0.1\%$ of total parameters). CLIP-Adapter~\cite{gao2025clipadapterbettervisionlanguagemodels} and DAC~\cite{gondal2024domain} use a bottleneck or a single linear layer. 
    \item \textbf{Preservation of Foundational Knowledge}: Majority of the methods agree that the pre-trained weights encase a more generalized understanding of the correlation between image and text modalities. These methods employ a frozen VLM backbone and utilize residual connections to ensure that the features extracted from pre-trained weights remain the dominant part of the final representation for downstream tasks. 
    \item \textbf{Rapid Convergence}: Modern research increasingly prioritizes quick adaptation for of VLMs for downstream tasks. State-of-the-art methods allow for instantaneous adaptation without the need for training. For instance, Tip-Adapter~\cite{zhang2021tipadaptertrainingfreeclipadapterbetter} and SuS-X~\cite{udandarao2023susxtrainingfreenameonlytransfer} introduce training-free frameworks that perform classification through cache retrieval. 
    \item \textbf{Maintaining Cross-Modal Alignment}: Contrastive learning aligns image and text features in higher-dimensional space. Recent methods have identified the need to maintain this alignment while addressing downstream tasks. For instance, MaPLe~\cite{khattak2023maplemultimodalpromptlearning} utilizes branch-aware coupling to ensure both encoders remain synchronised. Similarly, Domain Aligned CLIP (DAC)~\cite{gondal2024domain} optimises both intra-modal (image-to-image) and inter-modal (image-to-text) relationships.         

\end{enumerate}
By adhering to these four principles, the few-shot method using CLIP refines the high-dimensional manifold without compromising the inherent multimodal structure of the foundation models.

\textbf{Modality Gap}:  The existence of a "modality gap" between image and text features is a well-understood phenomenon in vision-language models \cite{modalitygap}. Recent studies \cite{modalitygap} have shown that image and text embeddings reside in two distinct and isolated conical regions within the high-dimensional feature space. Zero-shot classification using CLIP computes the dot product between these features and considers them as posteriors for class assignment. However, the range of these values is fundamentally constrained by the geometry of these two conic sections, which inherently creates ambiguity between corresponding image-text (positive) pairs and unmatched (negative) pairs. This limits the model's ability for classification.

\begin{figure}[ht]
    \centering
    \begin{subfigure}[b]{0.48\linewidth}
        \centering
        \includegraphics[width=\textwidth]{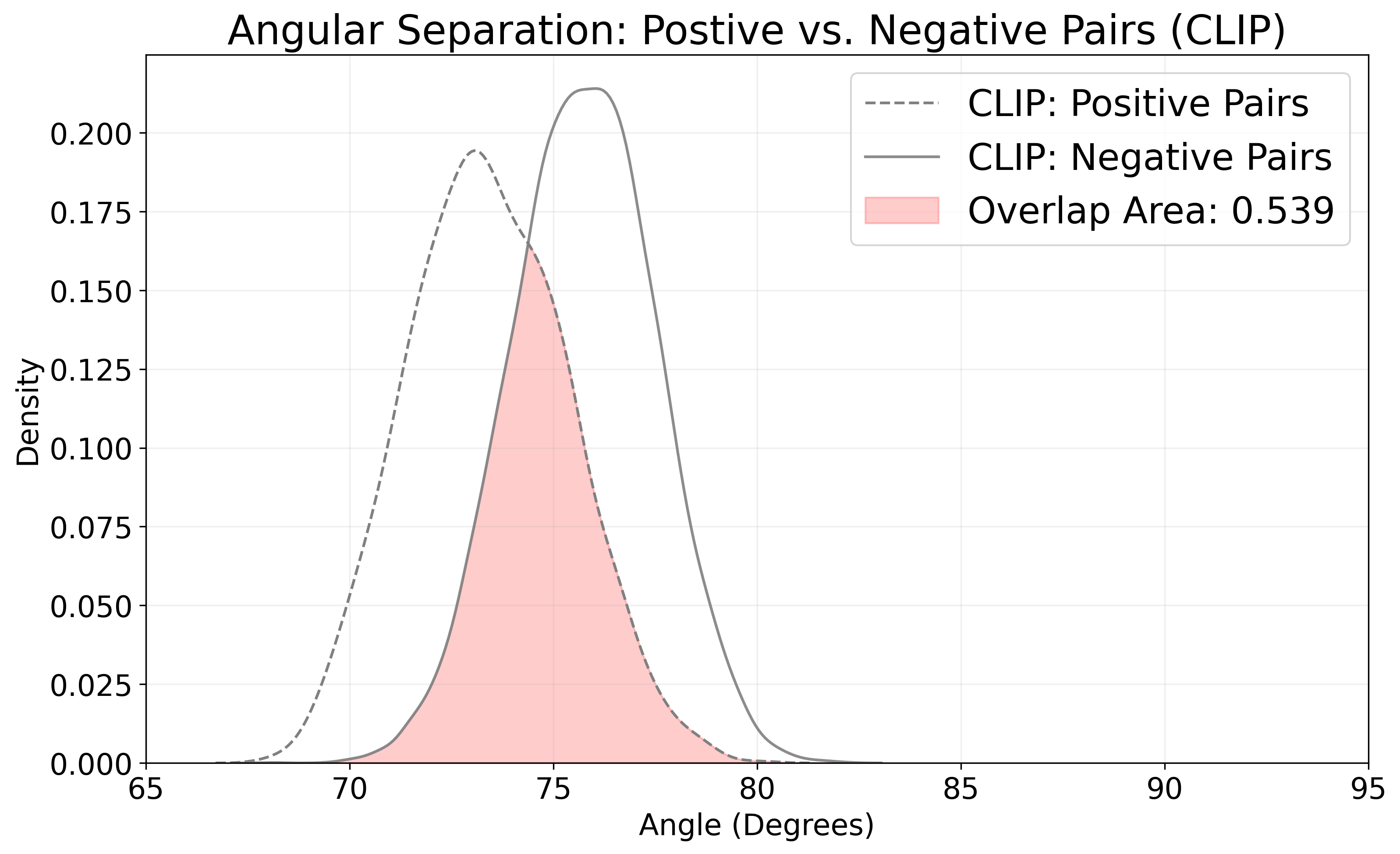}
        \caption{Zero-Shot (overlap: $0.539$)}
        \label{fig:ang_dist_zeroshot}
    \end{subfigure}    
    \hfill
    \begin{subfigure}[b]{0.48\linewidth}
        \centering
        \includegraphics[width=\textwidth]{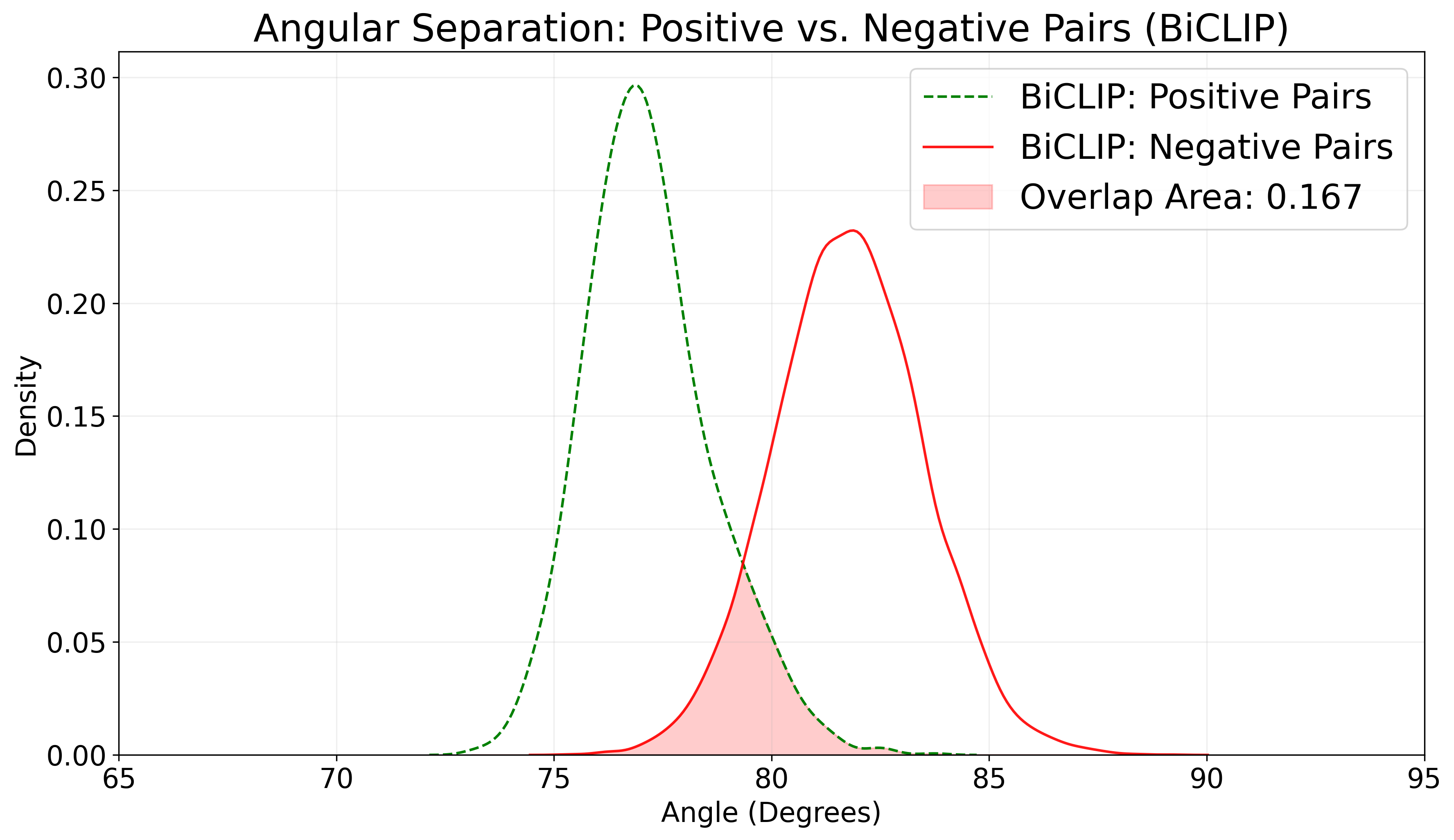}
        \caption{BiCLIP (Overlap: $0.167$)}
        \label{fig:dist_rbi}
    \end{subfigure}    
    \caption{Quantitative analysis of angular distribution on DTD~\cite{dtd} dataset. The zero-shot CLIP (a) shows significant overlap between positive and negative pairs. A structured geometric transformation on the image features (b) reduces the overlap area significantly.}
    \label{fig:manifold_comparison}
\end{figure}

This can be quantified by computing the overlap of the angular distribution of positive and negative pairs. Fig. \ref{fig:manifold_comparison} shows the probability distribution of angles computed between positive and negative image pairs on the DTD dataset~\cite{dtd}. DTD is a challenging fine-grained texture dataset with subtle semantic differences and complex visual patterns. Fig. \ref{fig:ang_dist_zeroshot} reveals that in the zero-shot CLIP, there is a significant overlap (area: $0.539$) in the positive and negative distributions. The model essentially fails to decouple matching pairs from the surrounding unmatched pairs, indicating that a simple dot product $(\cdot)$ is inadequate for reliable classification.


\textbf{Canonical Alignment}: Gupta \textit{et al.}~\cite{gupta2026canonicalizingmultimodalcontrastiverepresentation} theorize that multimodal models trained on independent web-scale data share a latent geometric structure, where the representations of different models—or different modalities within the same model—are related by a canonical transformation. Specifically, they suggest that the misalignment between image and text manifolds can be largely reconciled through orthogonal mappings. If the modality gap is fundamentally a problem of relative rotation and scaling between these manifolds, then a targeted rotation on individual modalities can achieve precise alignment in downstream tasks.

\textbf{Bilinear Adaptation of Contrastive VLMs}: We propose Bilinear CLIP (BiCLIP), a second-order lightweight architectural unit designed to explicitly bridge this gap through geometric transformation. Moving away from traditional additive residual adapters, we hypothesize that the optimal way to achieve the canonical alignment suggested by Gupta \textit{et al.}~\cite{gupta2026canonicalizingmultimodalcontrastiverepresentation} is by learning a weight matrix $W$ that performs a targeted geometric transformation of the high-dimensional manifold of the image and the text features.

The primary strength of BiCLIP lies in its extreme simplicity. It introduces a single multimodal interaction layer, allowing the model to widen the gap in angular distribution without compromising the integrity of CLIP’s pre-trained features. This design ensures that the model remains parameter and computationally efficient, requiring minimal training epochs for convergence, thus adhering to the principles of Few-Shot Adaptation. Fig. \ref{fig:manifold_comparison}(b) shows the angular distribution of positive and negative pairs after applying a learned transformation on the image feature. These results demonstrate that this simple geometric adjustment allows the model to explicitly mitigate confusion and enhance classification accuracy.

\textbf{Contributions}: We demonstrate that domain adaptation in VLMs can be framed as a geometric recovery problem. By utilizing a few-shot sample as anchors, BiCLIP recovers the underlying canonical transformations between disparate domains via a structured geometric transformation. 


The core contributions of this work are as follows:
\begin{enumerate}
    \item We extend multimodal canonicalization to domain shifts, hypothesizing that disparate domains are related by canonical geometric transformations that can be estimated via limited anchors.
    \item We introduce a simple bilinear unit that performs a non-destructive manifold transformation for better alignment.
    \item We provide a quantitative analysis of how BiCLIP reduces overlap of angular distributions in contrastive VLMs.
    \item We demonstrate SOTA or competitive performance across eleven standard benchmarks, including ImageNet~\cite{imagenet}, EuroSAT~\cite{eurosat}, and FGVC-Aircraft~\cite{fgvcaircraft}, proving the robustness and generalizability of BiCLIP.
\end{enumerate}

\section{Related Work}
\label{sec:related_work}

\subsection{Adaptation of Vision-Language Models}
Contrastive pre-trained VLMs like CLIP \cite{radford2021learningtransferablevisualmodels}, and SigLIP \cite{zhai2023sigmoidlosslanguageimage} have found immense success attributing their extensive training on web-scale datasets and showcase excellent classification capabilities out-of-the-box. However, these models underperform in specialized downstream tasks due to the domain gap between general web data and specific visual distributions (e.g., satellite imagery or fine-grained textures).

To bridge this gap, Parameter-Efficient Fine-Tuning (PEFT)~\cite{xu2023parameterefficientfinetuningmethodspretrained} has emerged as the standard paradigm, primarily categorized into: Prompt Learning-based~\cite{chen2023plotpromptlearningoptimal, huang2022unsupervisedpromptlearningvisionlanguage, zhou2022conditionalpromptlearningvisionlanguage, zhou2022conditionalpromptlearningvisionlanguage, shu2022testtimeprompttuningzeroshot, lu2022promptdistributionlearning},  and Adapter-based Methods~\cite{gao2025clipadapterbettervisionlanguagemodels, zhang2021tipadaptertrainingfreeclipadapterbetter}.

Zhou \textit{et al.} proposed CoOp \cite{Zhou_2022}, a prompt learning-based approach with a frozen CLIP backbone that optimizes learnable context vectors in the text encoder. Subsequent iterations like CoCoOp \cite{zhou2022conditionalpromptlearningvisionlanguage} and MaPLe \cite{khattak2023maplemultimodalpromptlearning} introduced input-conditional and multimodal prompting to improve generalization. However, these methods often require multiple extensive training strategies, complex architectural enhancements, and can be sensitive to hyperparameter tuning.

Adapter-based Methods such as CLIP-Adapter \cite{gao2025clipadapterbettervisionlanguagemodels} and Tip-Adapter \cite{zhang2021tipadaptertrainingfreeclipadapterbetter} introduce lightweight modules—typically bottleneck MLPs—into the frozen backbone. Tip-Adapter leverages a cache model of few-shot features to perform training-free adaptation. Our work diverges from these approaches and proposes a structured bilinear head that directly operates on the multimodal feature geometry. We achieve SOTA results with a significantly smaller parameter footprint and better preservation of the pre-trained semantic structure than traditional adapter-based methods.

\subsection{Geometric Representation and Multimodal Alignment}

Recent studies on the geometric properties of multimodal latent spaces have gained significant traction and allowed for documentation of the "modality gap", where image and text embeddings occupy disjoint regions of the hypersphere \cite{modalitygap}. This separation leads to suboptimal feature alignment for downstream tasks. Materzynska \textit{et al.}~\cite{materzynska2022disentanglingvisualwrittenconcepts} investigated the entanglement of concepts in image encoders, utilizing orthogonal projections to disentangle visual and written information. Similarly, Mistretta \textit{et al.}~\cite{mistretta2025crossgapexposingintramodal} proposed Modality Inversion (OVI) to bridge this gap by transforming intra-modal tasks into inter-modal ones, thereby enhancing alignment. In specialized domains, CP-CLIP maps embeddings to a unified "core-periphery" space to improve matching for medical zero-shot classification \cite{ricci2025boldsymbollambdaorthogonalityregularizationcompatiblerepresentation}.

While previous attempts have explored orthogonal constraints to maintain the semantic integrity of pre-trained spaces, these methods are often computationally intensive or require extensive structural adaptation. Our work builds on the foundational insight by Gupta \textit{et al.}~\cite{gupta2026canonicalizingmultimodalcontrastiverepresentation}, who suggested that independently trained multimodal manifolds are related by a shared orthogonal transformation—a principle known as multimodal canonicalization. 

We extend this theory to the domain-specific setting. Unlike existing methods, our approach utilizes an identity-initialized bilinear transformation to inherit the pre-trained capabilities and is constrained by an upper triangular structure to mitigate overfitting for learning task-specific geometric refinements. By analyzing the angular distribution and orthogonality deviation, we demonstrate that maintaining these geometric properties is essential for stable, high-performance few-shot classification
\section{Preliminaries}
\subsection{Contrastive Language-Image Pre-training Models (CLIP)}
\textbf{CLIP}~\cite{radford2021learningtransferablevisualmodels} is a dual-encoder architecture with an image encoder $f_i$ and a text encoder $f_t$. The encoders map the raw image $x$ and textual prompt $t$ into a common embedding space. CLIP model are extensively trained on large-scale datasets~\cite {radford2021learningtransferablevisualmodels} using contrastive loss such that matching pairs of image-text are closer to each other and unmatched pairs are farther away. CLIP exhibits excellent zero-classification capabilities due to this contrastive training. 

During inference, image features are extracted as $i=f_i(x)$, where $i \in R^{1XD}$, where $D$ is the dimensionality of the embedding space. In a classification scenario with $K$ classes, text features $\{\mathbf{t}_k\}_{k=1}^K$ are extracted as $t_k = f_t(prompt_k)$, where $t_k \in R^{1XD}$. Typically both the features are first normalized such that $||i|| = ||t|| = 1$. Then the posterior of a class $k$ is calculated as the softmax of the cosine similarity between $i$ and $t_k$.

\begin{equation}
P(y=k | \mathbf{x}) = \frac{\exp(e^{s} \cdot \cos(\mathbf{i}, \mathbf{t}_k))}{\sum_{j=1}^K \exp(e^{s} \cdot \cos(\mathbf{i}, \mathbf{t}_j))}
\end{equation}
where $e^s$ is the logit scale (with $s$ being the learnable parameter), and both $\mathbf{i}$ and $\mathbf{t}$ are $\ell_2$-normalized such that $\cos(\mathbf{i}, \mathbf{t}_k) = \mathbf{i} \mathbf{t}_k^\top$.

\subsection{Sigmoid Loss for Language-Image Pre-training (SigLIP)}
SigLIP~\cite{zhai2023sigmoidlosslanguageimage} treats each image-text pair as an independent binary classification task. Given a batch of $N$ image features $\{\mathbf{i}_n\}_{n=1}^N$ and $M$ text features $\{\mathbf{t}_n\}_{n=1}^M$, the similarity score for any pair $(j, k)$ is defined as:
\begin{equation}
s_{j,k} = e^s \cdot (\mathbf{i}_j\mathbf{t}_k^\top) + b
\end{equation}

where $e^s$ is the logit scale and $b$ is a learnable bias. The training objective minimizes the binary cross-entropy loss.

During zero-shot inference, while the training objective is sigmoid-based, the posterior probability that image $i_j$ belongs to class $k$ is typically computed using a softmax over the similarity scores of all $K$ candidate classes to maintain consistency with the classification task:

\begin{equation}
P(y=k | \mathbf{x}) = \frac{\exp(s_{\mathbf{j}, k})}{\sum_{l=1}^K \exp(s_{\mathbf{j}, l})}
\end{equation}

While SigLIP achieves a stable embedding geometry at scale, like CLIP, it still exhibits a characteristic modality gap. BiCLIP seeks to canonicalize this space by learning a structured transformation to align these disparate domains.

\subsection{Shortcoming of the zero-shot CLIP} In CLIP and SigLIP, probabilities are derived through a simple dot product between fixed image and text embeddings. In zero-shot CLIP, the relationship between $i$ and $t_k$ is determined by the pre-trained weights. As we demonstrated in Section~\ref{sec:intro}, the angular distribution on the DTD dataset shows a significant overlap (area: $0.539$), which severely degrades classification performance. 

This overlap suggests that the features are constrained to a manifold that fails to account for the semantic shifts inherent in specialized downstream tasks. This necessitates an adaptation strategy that provides the flexibility to "warp" or "align" these features. Such a mechanism would effectively canonicalize the domain-specific space, narrowing the angular distribution of positive pairs and reducing the manifold overlap.
\section{BiCLIP: Structured Bilinear Alignment}
To overcome the inherent shortcomings of zero-shot contrastive VLMs, we propose BiCLIP. Our core hypothesis is that the modality gap is a rotational misalignment that can be resolved through a targeted geometric transformation that allows image features to be dynamically "rotated" and "canonicalized" into alignment with textual anchors.

\subsection{BiCLIP Theory}
The motivation for BiCLIP is to apply a learnable geometric transformation to the image feature vector before it interacts with the text embeddings. Let $\mathbf{i_j} \in \mathbb{R}^{1 \times D}$  be the $j^{th}$ image features and $\mathbf{t_k} \in \mathbb{R}^{1 \times D}$ be the $k^{th}$ text features. Instead of a direct dot product, we first transform the image feature using a weight matrix $\mathbf{W} \in \mathbb{R}^{D \times D}$, resulting in a transformed feature $\mathbf{i_j}^{'} = \mathbf{i_j} \mathbf{W}$. The similarity score $s^{bi}$ is then computed as the dot product between this transformed image feature and the textual features as:

\begin{equation}
s^{bi}_{j,k} = \mathbf{i_j}^{'} \cdot \mathbf{t_k}^\top = (\mathbf{i_j} \mathbf{W}) \mathbf{t_k}^\top
\end{equation}

By the associative property, this is equivalent to the bilinear form:
\begin{equation}
S(\mathbf{i}, \mathbf{t}) = \mathbf{i} \mathbf{W} \mathbf{t}^\top
\end{equation}

This equivalence demonstrates that cross-modal bilinear interaction is, in effect, a learnable alignment operator. By optimizing $\mathbf{W}$, the network learns to align the feature space, narrowing the modality gap by orienting the image features toward their corresponding targets in the latent space.

\subsection{Upper Triangular Structured Constraint}
A significant challenge in few-shot learning, especially in high-dimensional spaces, is the risk of overfitting. The $\mathbf{W}$ matrix contains $D^2$ parameters; for SigLIP with 768-dimensional embedding, this results in $5.9 \times 10^5$ trainable weights. To mitigate the risk of manifold collapse, we impose a structural constraint by restricting $\mathbf{W}$ to be an upper triangular matrix. 
This serves as a regularizer in two ways:

\textbf{Hierarchical Dependence}: The constraint ensures that each dimension of the transformed feature $\mathbf{i^{'}}_{j}$ is a linear combination of its original value and only the subsequent dimensions. 

\textbf{Parameter Reduction}: This constraint reduces the total number of trainable parameters by nearly half, to $\frac{D(D+1)}{2}$, mitigating overfitting. In the context of BiCLIP, this prevents the matrix from performing extreme non-rigid warping that would otherwise displace the foundational knowledge of the frozen backbone. 
This form of structural regularization is inspired by the Cholesky decompositionlearning~\cite{Pourahmadi_2011} and studies in sparse matrix. 


In this work, we frequently utilize the term "rotation" to describe the geometric effect of $\mathbf{W}$ on the image manifold. We clarify that this is not a rigid rotation in the sense of an orthogonal matrix ($R^\top R = \mathbf{I}$). Instead, our use of an Upper Triangular constraint is a choice of Geometric Canonicalization. To be precise $\mathbf{W}$ performs a non-rigid geometric transformation that aligns the modalities, its primary role is the canonicalization and alignment of the target domain feature space.

\subsection{BiCLIP: Bilinear Adaptation for CLIP}
To integrate BiCLIP into modern vision-language frameworks, we substitute the standard dot-product similarity with the bilinear term. This adaptation is agnostic to the underlying objective function, allowing BiCLIP to enhance both symmetric softmax and pairwise sigmoid architectures.

\begin{figure}[t]
    \centering
    \includegraphics[width=\linewidth]{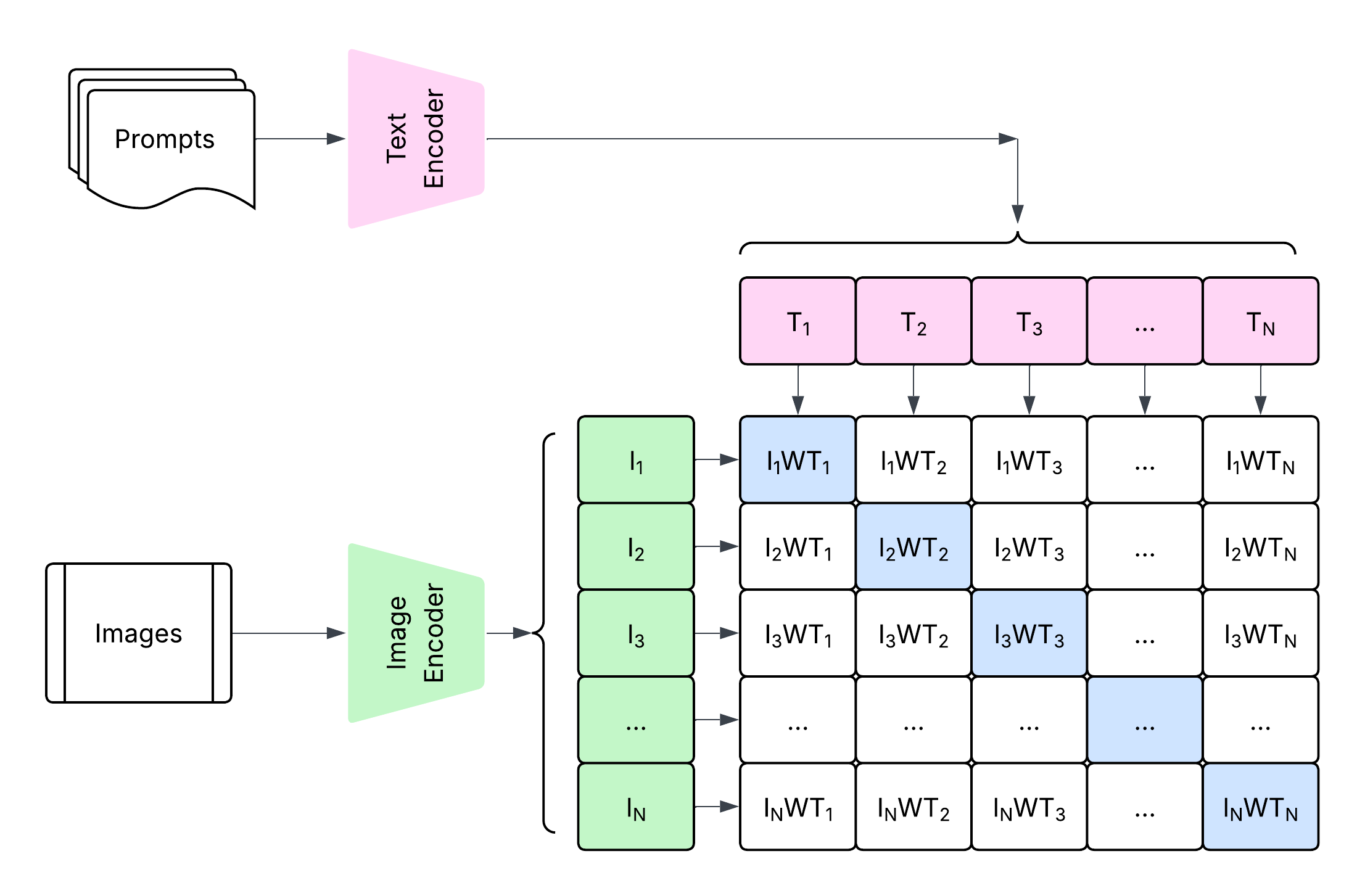}
    \caption{\textbf{The BiCLIP Adaptation Framework.} Unlike standard CLIP which relies on a fixed dot product, BiCLIP introduces a trainable, structured transformation matrix $\mathbf{W}$ between the image and text modalities. }
    \label{fig:biclip_architecture}
\end{figure}

In the standard CLIP framework, the model is trained to maximize the cosine similarity between positive image-text pairs in a batch while minimizing it for negative combinations. By introducing the transformation weight matrix $\mathbf{W}$, the similarity score 
for an image feature $\mathbf{i}_j$ and text feature $\mathbf{t}_k$ is computed as:$S^{\text{bi}}_{j,k} = e^s \cdot (\mathbf{i}_j \mathbf{W} \mathbf{t}_k^\top)$. As illustrated in Figure~\ref{fig:biclip_architecture}, we transition from the standard dot-product similarity used in CLIP to a learnable bilinear formulation.

We maintain the symmetric cross-entropy loss, which treats the classification as both an image-to-text and a text-to-image retrieval task. For a batch of $N$ pairs, the loss is defined as:

\begin{equation}
\begin{aligned}
\mathcal{L}_{\text{BiCLIP}} = -\frac{1}{2N} \sum_{n=1}^{N} \Big[ 
    &\log \frac{\exp(S^{\text{bi}}_{n,n})}{\sum_{j=1}^N \exp(S^{\text{bi}}_{n,j})} \\ 
    + &\log \frac{\exp(S^{\text{bi}}_{n,n})}{\sum_{j=1}^N \exp(S^{\text{bi}}_{j,n})} \Big]
\end{aligned}
\end{equation}

By using $S^{\text{bi}}_{n,j}$ in place of the dot product, the softmax competition forces $\mathbf{W}$ to learn a transformation that pushes the image and text pairs into alignment.

\subsection{BiSigLIP: Bilinear Adaptation for SigLIP}
The adaptation for SigLIP involves embedding the bilinear transformation directly into the sigmoid logit calculation. The modified similarity score $S^{\text{bi}}_{j,k}$ incorporates the SigLIP-specific learnable bias $b$:

\begin{equation}
S^{\text{bi}}_{j,k} = e^s \cdot (\mathbf{i}_j \mathbf{W} \mathbf{t}_k^\top) + b
\end{equation}

where $e^s$ is the logit scale. The model is then optimized using the pairwise binary cross-entropy (sigmoid) loss:
\begin{equation}
\mathcal{L}_{\text{BiSigLIP}} = - \frac{1}{N} \sum_{j,k} \log \sigma (y_{j,k} \cdot S^{\text{bi}}_{j,k})
\end{equation}

where $y_{j,k} = 1$ for positive pairs ($j=k$) and $y_{j,k} = -1$ for negative pairs ($j \neq k$). Because the sigmoid loss treats each pair as an independent binary classification task, the bilinear matrix $\mathbf{W}$ can more precisely target the domain-specific modality gap for each specific class. This facilitates a robust alignment of the multimodal features for specialized domains.


\section{Experimental Methodology}
\label{sec:methodology}

\subsection{Datasets}
We evaluate BiCLIP across the standard few-shot image recognition datasets\cite{Zhou_2022}. These include: \textbf{ImageNet}~\cite{imagenet} and \textbf{Caltech101}~\cite{caltech101} for generic objects; \textbf{OxfordPets}~\cite{oxfordpets}, \textbf{StanfordCars}~\cite{stanfordcars}, \textbf{Flowers102}~\cite{flowers102}, and \textbf{FGVCAircraft}~\cite{fgvcaircraft} for fine-grained classification; \textbf{SUN397}~\cite{sun397} for scene recognition; \textbf{DTD}~\cite{dtd} for texture analysis; \textbf{EuroSAT}~\cite{eurosat} for satellite imagery; \textbf{UCF101}~\cite{ucf101} for action recognition; and \textbf{Food101}~\cite{food101} for food classification. For each dataset, we follow the standard few-shot evaluation protocol, using 1, 2, 4, 8, and 16 shots for training and evaluating on the full test sets.

\subsection{Implementation Details}
All experiments are conducted using \textbf{CLIP (ViT-B/16)} and \textbf{SigLIP (ViT-B/16)}. SigLIP is trained on the \textbf{WebLI} dataset~\cite{chen2022pali}. Both models share the same Vision Transformer (ViT) architecture. However, SigLIP (ViT-B/16) produces richer high-dimensional features ($D=768$) compared to the OpenAI's CLIP ($D=512$) model.

SigLIP embeddings are generally more robust and demonstrate superior zero-shot performance, a result of the Sigmoid loss and the massive scale of the WebLI dataset. By evaluating BiCLIP on both backbones, we demonstrate the generalizability of BiCLIP across varying feature dimensionalities and pre-training objectives.

We utilize the AdamW optimizer with a weight decay of $0.1$ and an initial learning rate of $10^{-4}$ for training. All experiments are conducted on a NVIDIA 2080Ti GPU. Depending on the complexity and size of the dataset, we train for a range of 20 to 50 epochs to ensure convergence of the bilinear transformation matrix $\mathbf{W}$.

\subsection{Identity Initialization}
To preserve the zero-shot capabilities of the pre-trained backbones, we initialize the transformation matrix $\mathbf{W}$ as an Identity matrix $\mathbf{I} \in \mathbb{R}^{D \times D}$. Let $\mathbf{X} \in \mathbb{R}^{B \times D}$ represent a batch of $B$ image features and $\mathbf{T} \in \mathbb{R}^{C \times D}$ denote the text features for $C$ classes. The similarity score in the zero-shot setting is computed as $\mathbf{S} = \mathbf{X}\mathbf{T}^\top$. For BiCLIP versions, the score is computed as $\mathbf{S}^{\text{bi}} = (\mathbf{X}\mathbf{W})\mathbf{T}^\top$. 

Under an identity initialization, the similarity score ($\mathbf{X}\mathbf{I}\mathbf{T}^\top = \mathbf{X}\mathbf{T}^\top$) simplifies to the zero-shot score. This ensures that the model's performance is identical to the zero-shot baseline at the onset of training and provides a robust initialization. 






\section{Experimental Results}
\label{sec: experiemtal_results}
We present results in three main subsections: overall performance in a 16-shot setting, comparison to state-of-the-art in standard (1, 2, 4, 8, and 16 shot) settings, and analysis of the geometric and structural properties of the transformation matrix $W$.

\subsection{Main Results: 16-Shot Performance}
First, we present a detailed comparison of our proposed bilinear adaptation methods against their respective zero-shot baselines. Table~\ref{tab:main_results} summarizes the performance across all datasets using 16 training shots per class.

\begin{table*}[th]
\small
\centering
\begin{tabular}{l | ccc | ccc}
\toprule
\textbf{Dataset} & \textbf{Zero-Shot} & \textbf{BiCLIP} & \textbf{Delta} & \textbf{Zero-Shot} & \textbf{BiSigLIP} & \textbf{Delta} \\
 &  \textbf{CLIP} & \textbf{(Ours)} & \textbf{$\Delta$} & \textbf{SigLIP} & \textbf{(Ours)} & \textbf{$\Delta$} \\
\midrule
ImageNet & 68.84 & \textbf{71.69} & +2.85 & 74.89 & \textbf{76.73} & +1.83 \\
DTD & 42.82 & \textbf{71.01} & +28.19 & 62.23 & \textbf{73.94} & +11.70 \\
EuroSAT & 48.22 & \textbf{85.13} & +36.91 & 35.35 & \textbf{77.50} & +42.15 \\
Flowers102 & 70.99 & \textbf{94.97} & +23.99 & 81.15 & \textbf{96.11} & +14.96 \\
FGVCAircraft & 24.60 & \textbf{45.21} & +20.61 & 45.99 & \textbf{49.41} & +3.42 \\
OxfordPets & 89.04 & \textbf{93.30} & +4.24 & 92.31 & \textbf{92.80} & +0.49 \\
Food101 & 88.73 & \textbf{90.09} & +1.36 & 92.19 & \textbf{92.33} & +0.14 \\
Caltech101 & 89.93 & \textbf{93.97} & +4.04 & 95.23 & \textbf{97.06} & +1.83 \\
SUN397 & 63.50 & \textbf{74.27} & +10.77 & 65.85 & \textbf{74.24} & +8.38 \\
UCF101 & 68.07 & \textbf{82.95} & +14.88 & 71.50 & \textbf{78.85} & +7.35 \\
StanfordCars & 63.71 & \textbf{82.63} & +18.92 & 88.81 & \textbf{92.12} & +3.31 \\
\midrule
\textbf{Average} & 65.31 & \textbf{80.47} & \textbf{+15.16} & 73.22 & \textbf{81.91} & \textbf{+8.69} \\
\bottomrule
\end{tabular}
\caption{Main Results: 16-Shot Performance Comparison. We report Top-1 Accuracy (\%) for Zero-Shot baselines and our proposed Bilinear adaptation (BiCLIP and BiSigLIP). $\Delta$ represents the gain over the respective zero-shot baseline.}
\label{tab:main_results}
\end{table*}

\textbf{Performance Gain over Baselines:} Bilinear adaptation of CLIP and SigLIP shows notable and consistent improvement across all the datasets. BiCLIP achieves an average accuracy of $80.47\%$, marking a substantial $+15.16\%$ absolute improvement over the zero-shot baseline ($65.31\%$). Similarly, BiSigLIP pushes the already strong SigLIP baseline from $73.22\%$ to $81.91\%$, a gain of $+8.69\%$. These results confirm that a learnable geometric transformation of the image manifold is highly effective for aligning pre-trained multimodal features.

\textbf{Adaptability to Fine-Grained Tasks}: Zero-shot models struggle to generalize in domains where the images differ significantly from general web-scale pre-training data. BiCLIP and BiSigLIP methods demonstrate a particular aptitude in these specialized fine-grained tasks. They show significant improvements of $+36.91\%$ and $+42.15\%$ on the Eurosat (satellite imagery classification). Similar performance improvements are observed on Flowers102, FGVCAircraft, and the DTD dataset. These results suggest that BiCLIP and BiSigLIP capture the intra-class features required for fine-grained recognition.

\textbf{Generalizability}: It is of significance that bilinear adaptation remains consistent across both CLIP and SigLIP. Even on datasets where zero-shot performance is already high, such as Caltech101 and StanfordCars, Bilinear adaptation manages to refine the feature space further, yielding consistent improvements. This suggests that bilinear adaptation is agnostic to the VLM backbone and shows generalization characteristics.

\subsection{Comparison to SOTA: Few-Shot Performance Analysis}
\begin{figure*}[t]
    \centering
    
    \begin{subfigure}[b]{0.24\textwidth}
        \centering
        \includegraphics[width=\textwidth]{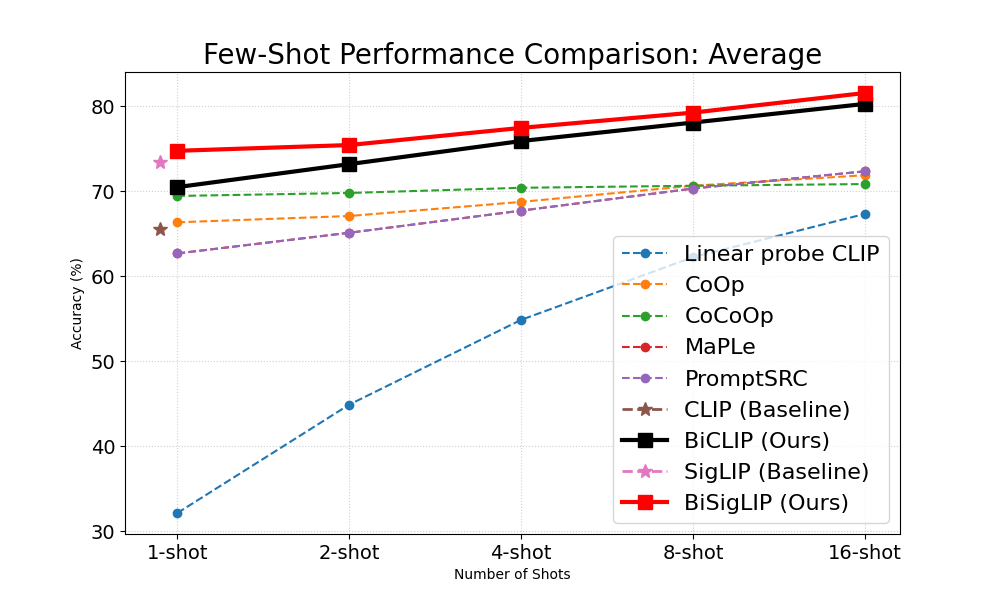}
        \caption{Average}        
    \end{subfigure}
    \hfill
    \begin{subfigure}[b]{0.24\textwidth}
        \centering
        \includegraphics[width=\textwidth]{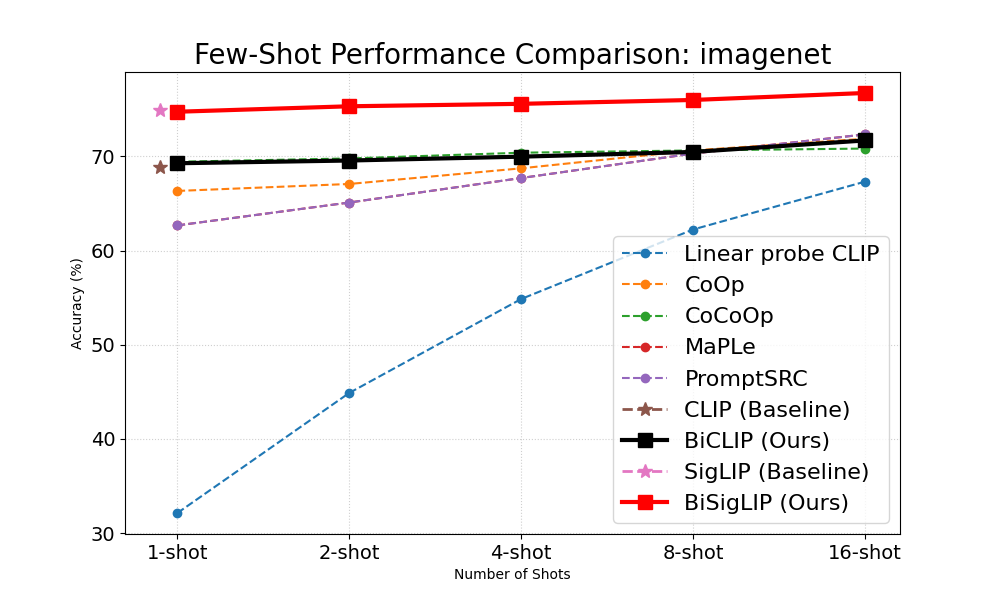}
        \caption{Imagenet}
    \end{subfigure}
    \hfill
    \begin{subfigure}[b]{0.24\textwidth}
        \centering
        \includegraphics[width=\textwidth]{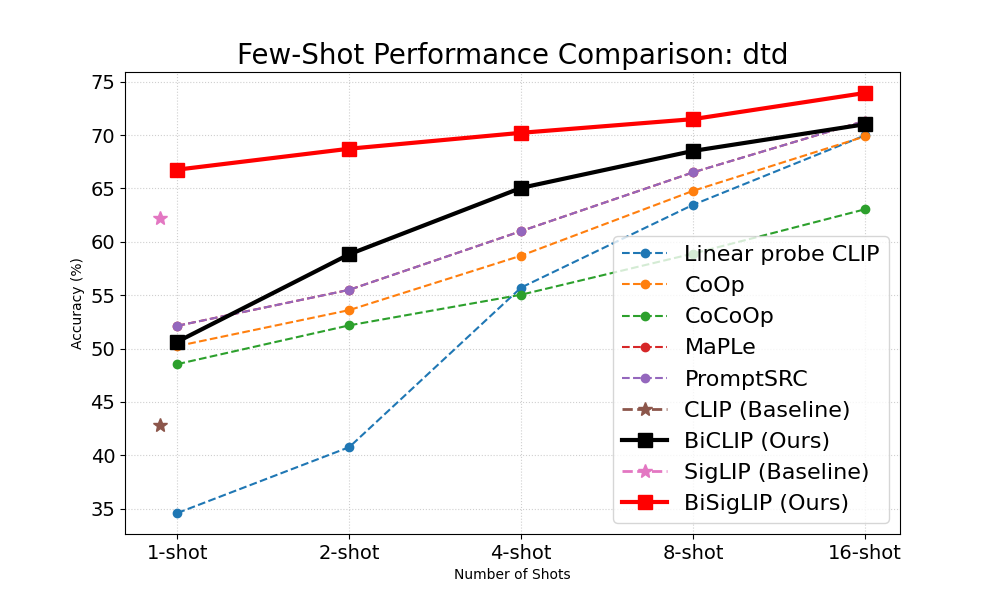}
        \caption{DtD}
    \end{subfigure}
    \hfill
    \begin{subfigure}[b]{0.24\textwidth}
        \centering
        \includegraphics[width=\textwidth]{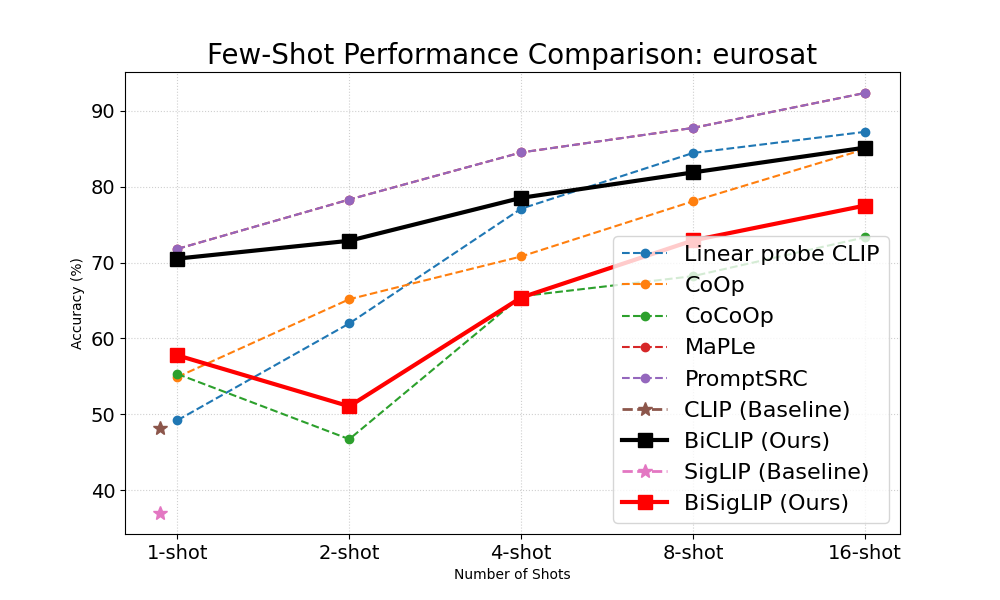}
        \caption{Eurosat}
    \end{subfigure}    
   
    \begin{subfigure}[b]{0.24\textwidth}
        \centering
        \includegraphics[width=\textwidth]{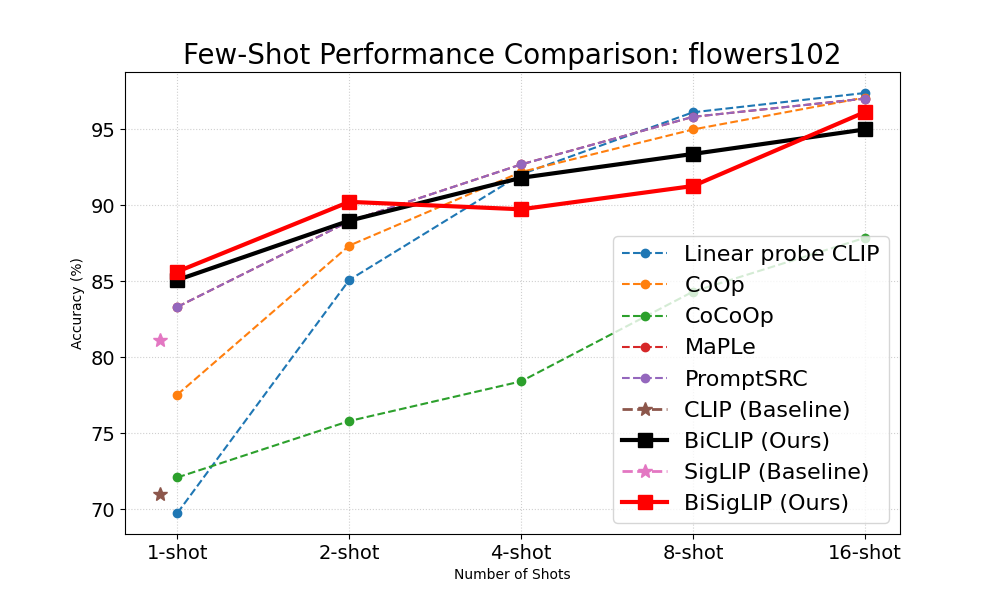}
        \caption{Flowers102}
    \end{subfigure}
    \hfill
    \begin{subfigure}[b]{0.24\textwidth}
        \centering
        \includegraphics[width=\textwidth]{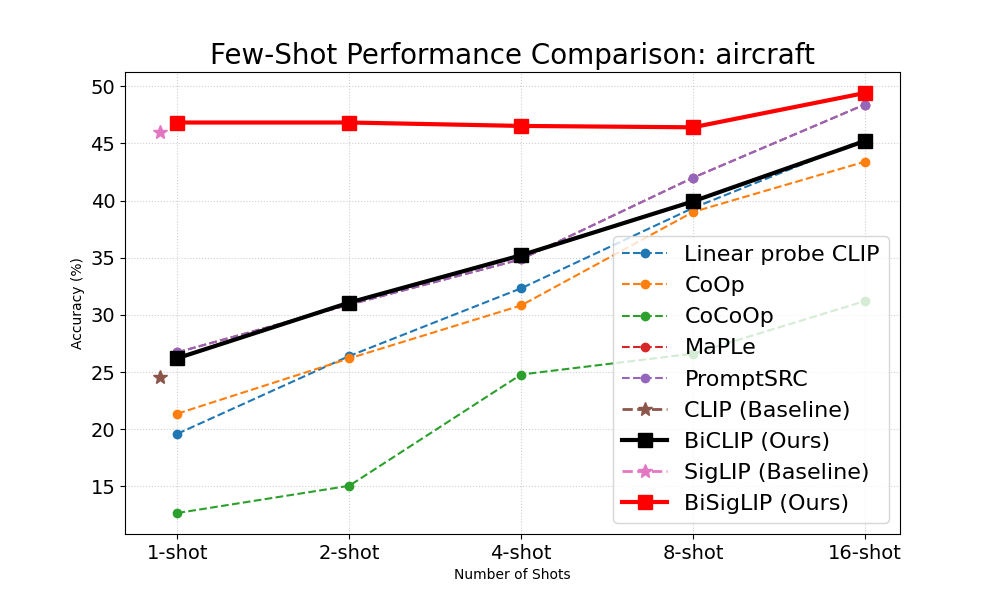}
        \caption{FGVCAircraft}
    \end{subfigure}
    \hfill
    \begin{subfigure}[b]{0.24\textwidth}
        \centering
        \includegraphics[width=\textwidth]{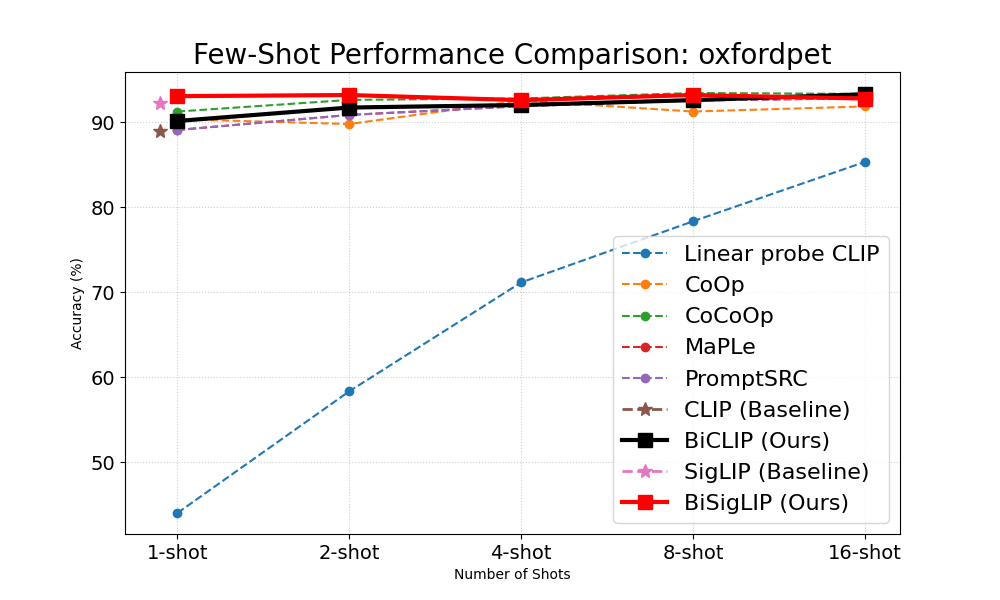}
        \caption{Oxfordpets}
    \end{subfigure}
    \hfill
    \begin{subfigure}[b]{0.24\textwidth}
        \centering
        \includegraphics[width=\textwidth]{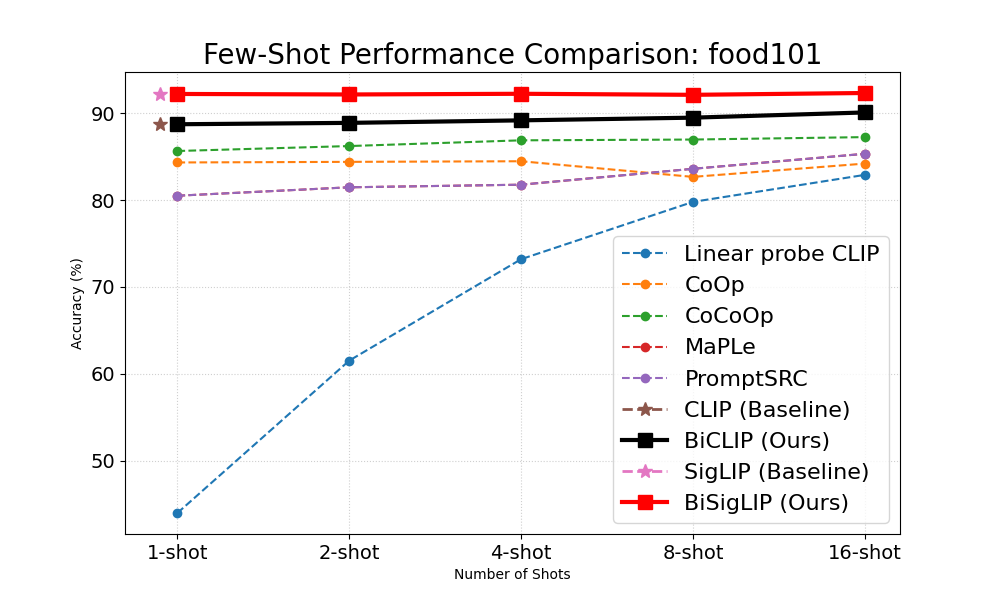}
        \caption{Food101}
    \end{subfigure}

    \begin{subfigure}[b]{0.24\textwidth}
        \centering
        \includegraphics[width=\textwidth]{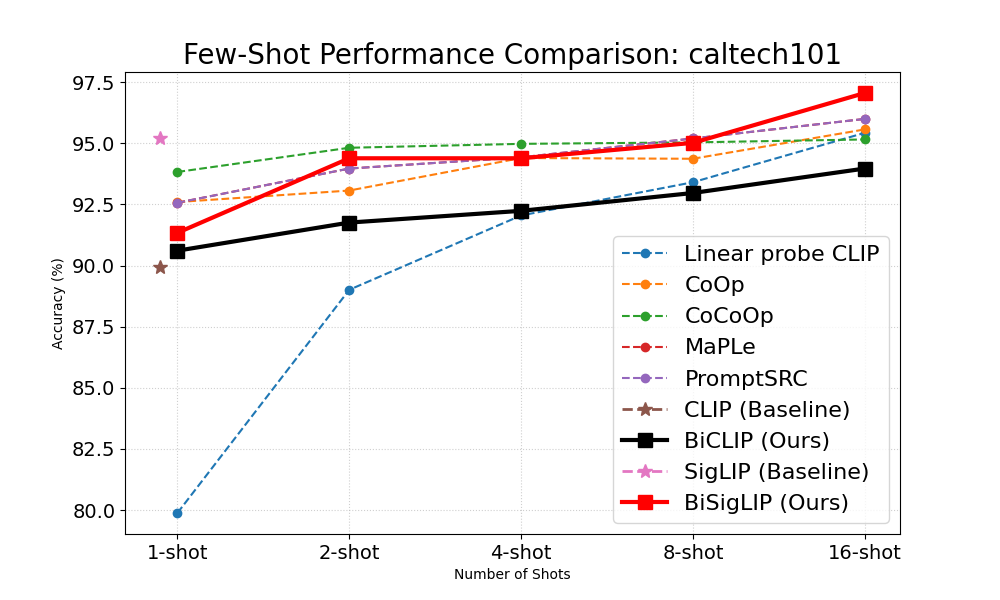}
        \caption{Caltech101}
    \end{subfigure}
    \hfill
    \begin{subfigure}[b]{0.24\textwidth}
        \centering
        \includegraphics[width=\textwidth]{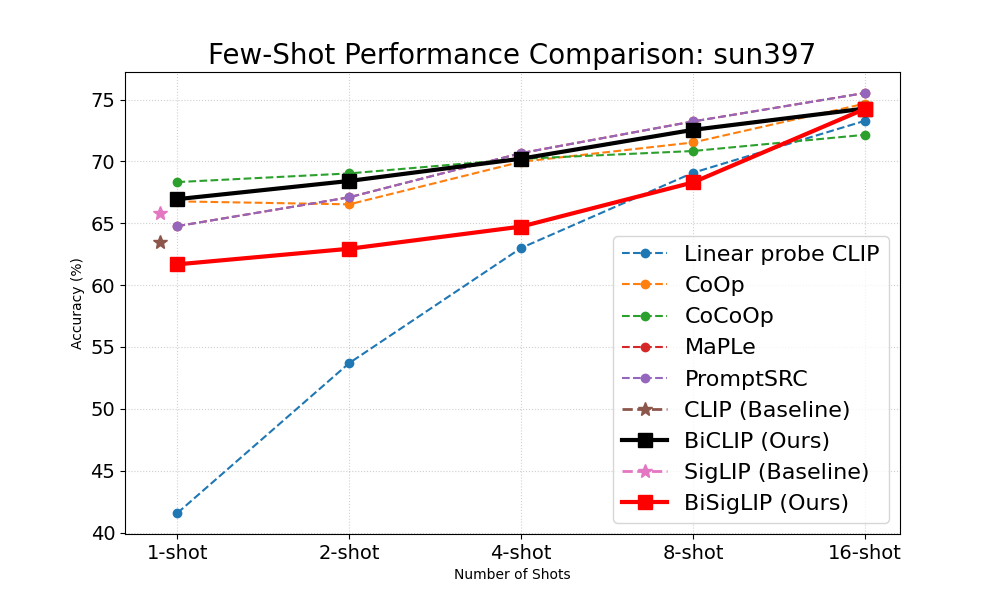}
        \caption{Sun397}
    \end{subfigure}
    \hfill
    \begin{subfigure}[b]{0.24\textwidth}
        \centering
        \includegraphics[width=\textwidth]{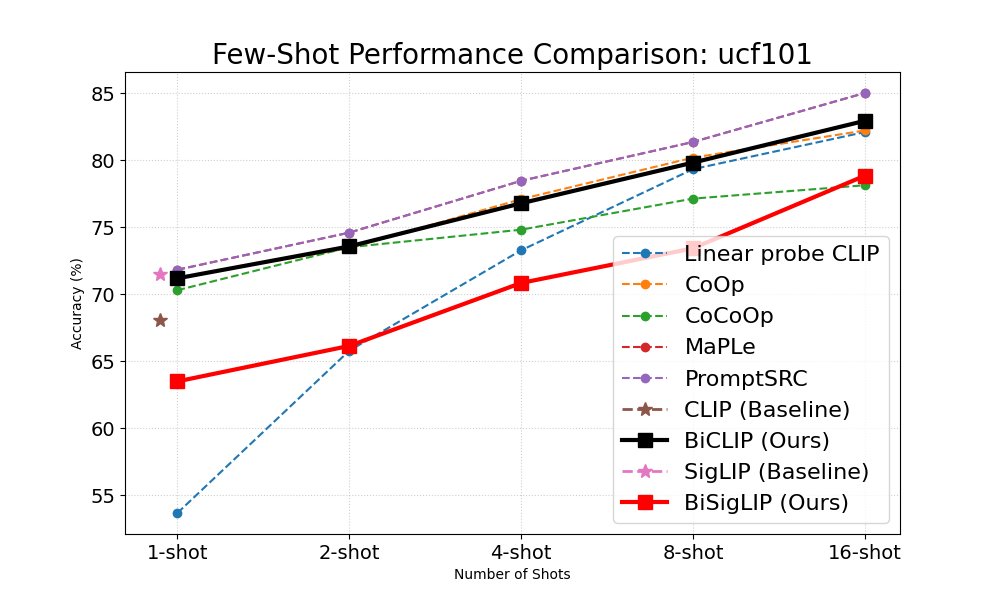}
        \caption{UCF101}
    \end{subfigure}
    \hfill
    \begin{subfigure}[b]{0.24\textwidth}
        \centering
        \includegraphics[width=\textwidth]{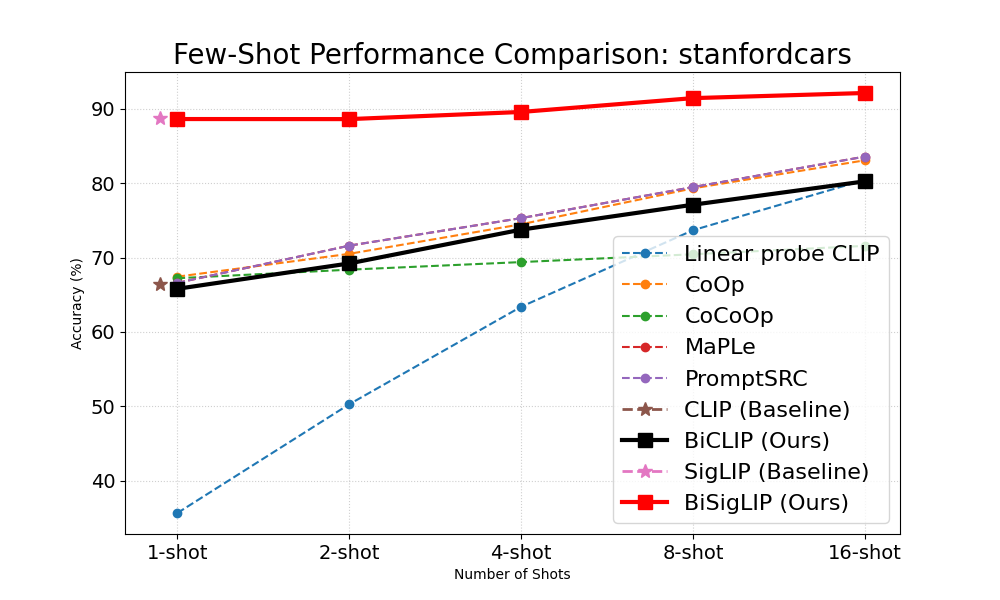}
        \caption{Stanfordcars}
    \end{subfigure}
    
    \caption{Few-shot performance comparison on various datasets. Our methods BiCLIP (black) and BiSigLIP (red) significantly outperform existing prompt tuning baselines across 1, 2, 4, 8, and 16 shots.}
    \label{fig:few_shot}
\end{figure*}

We conduct experiments across the standard 1, 2, 4, 8, and 16 shots settings. Figure~\ref{fig:few_shot} illustrates the performance curves of BiCLIP and BiSigLIP compared to five state-of-the-art baselines, including classic Linear Probe adaptation methods, prompt tuning variants CoOp \cite{Zhou_2022} and CoCoOp \cite{zhou2022conditionalpromptlearningvisionlanguage}, and more recent multimodal prompt learning techniques like MaPLe \cite{khattak2023maplemultimodalpromptlearning} and PromptSRC \cite{khattak2023selfregulatingpromptsfoundationalmodel}.

\textbf{Performance over 1,2-shot settings}: BiCLIP and BiSigLIP maintain a predictable high performance in low 1,2-shot settings. Fig. \ref{fig:few_shot} (a) shows the average scores over all datasets, where BiCLIP and BiSigLIP outperform in 1-shot and 2-shot settings. This can be attributed to the identity initialization of the transformation matrix that enables the model to start with an optimal zero-shot performance. Prompt-based methods like CoOp and MaPLe often require more samples for stable training and struggle in 1 and 2-shot regimes. 

\textbf{Simplicity and efficiency}: Bilinear adaptation offers a simple and efficient approach by applying a single matrix multiplication directly in the latent space. State-of-the-art methods such as MaPLe \cite{khattak2023maplemultimodalpromptlearning} and CoCoOp \cite{zhou2022conditionalpromptlearningvisionlanguage} require intensive training, complex optimization schedules, and are highly sensitive to the initialization choices. 
Bilinear adaptation adheres to the few-shot adaptation principles as it incorporates a simple transformation $W$ matrix with a minimal parameter footprint and requires limited training cycles. 


\textbf{Consistency across domains}: BiCLIP shows resilience to domain shifts. On complex datasets such as EuroSAT, DTD, and FGVCAircraft, it shows a consistent performance trajectory. This underscores the effectiveness of geometric feature alignment, which can capture the visual discriminative cues required for specialized domains.

\subsection{Analysis and Ablation}

\textbf{Angular Distribution}:
To better understand the effectiveness of the BiCLIP, we analyze the angular distribution between positive and negative image-text pairs. The overlap between the angular distributions is indicative of better alignment of matching image-text pairs for better classification. We first extract $L_2$-normalized image and text features to calculate the angular distance in degrees via $\theta = \arccos\left( \frac{\mathbf{i} \cdot \mathbf{t}}{\|\mathbf{i}\| \|\mathbf{t}\|} \right)$. We compute the angle between all positive pairs, and randomly sample n ($n=5$) negative samples per image in the test set. We estimate continuous probability density functions, $p_{pos}(\theta)$ and $p_{neg}(\theta)$ using Kernel Density Estimation (KDE). The overlap is then calculated as the area under the curve using Simpson’s Rule numerical integration: $\int \min(p_{pos}(\theta), p_{neg}(\theta)) \, d\theta$. A high overlap area indicates a congested latent space where correct and incorrect classes are geometrically indistinguishable, while a reduction signifies a more discriminative and well-aligned image and text features.

        

As illustrated in Fig. \ref{fig:manifold_comparison}(a) for the DTD dataset, zero-shot CLIP suffers from significant overlap (area: $0.539$), where the distributions for positive and negative pairs are largely indistinguishable. Bilinear adaptation effectively applies a targeted transformation on the image features, thus decreasing the overlap in the latent space. As shown in Fig. \ref{fig:manifold_comparison}(b), BiCLIP strategically shifts and narrows these distributions, reducing the overlap area to $0.167$ on DTD. As shown in Table \ref{tab:angular_dist}, this trend is observed across all datasets; BiCLIP reduces the average overlap from $0.209$ to $0.077$.

\begin{table}[th]
\centering
\begin{tabular}{lccc}
\toprule
Dataset & CLIP & \textbf{BiCLIP} & \textbf{$\Delta$ (Reduction)} \\
\midrule
FGVCAircraft & 0.327 & 0.175 & 0.152 \\
OxfordPets   & 0.100 & 0.044 & 0.056 \\
Flowers102   & 0.181 & 0.026 & 0.155 \\
StanfordCars & 0.071 & 0.048 & 0.023 \\
Food101      & 0.056 & 0.041 & 0.015 \\
DTD          & 0.539 & 0.167 & 0.372 \\
EuroSAT      & 0.596 & 0.187 & 0.409 \\
SUN397       & 0.127 & 0.037 & 0.090 \\
Caltech101   & 0.073 & 0.016 & 0.057 \\
UCF101       & 0.161 & 0.062 & 0.099 \\
ImageNet     & 0.068 & 0.039 & 0.029 \\
\midrule
\textbf{Average} & \textbf{0.209} & \textbf{0.077} & \textbf{0.133} \\
\bottomrule
\end{tabular}
\caption{Comparison of overlap in average angular distributions between image and text embeddings for zero-shot CLIP and BiCLIP. Lower values indicate better separation.}
\label{tab:angular_dist}
\end{table}

\textbf{Orthogonality of the $W$ matrix}:
Recent analysis of contrastive VLM models by Gupta \textit{et al.}~\cite{gupta2026canonicalizingmultimodalcontrastiverepresentation} demonstrate that independently trained contrastive models are related by a shared orthogonal map. This suggests that the underlying semantic structure across modalities is preserved through rotations. Inspired by this, we compute the normalized Frobenius norm $\|\mathbf{W}^\top \mathbf{W} - \mathbf{I}\|_F / D$ of the trained rotation matrix $\mathbf{W}$.

Tab. \ref{tab:orthogonality_analysis} shows the orthogonal error for all datasets. Our analysis confirms that $\mathbf{W}$ maintains orthogonality after convergence. On datasets like ImageNet (0.009) and Food101 (0.006), the normalized error is nearly negligible, indicating that the zero-shot manifold is already near the canonical state. However, on fine-grained datasets, we observe a slight departure from pure orthogonality. For instance, EuroSAT (0.024) and DTD (0.055) seem to require more non-rigid transformation. These finding aligns with recent work on multimodal canonicalization \cite{gupta2026canonicalizingmultimodalcontrastiverepresentation}, which shows that independently trained contrastive models are related by a global orthogonal map. We extend this notion to the domain-specific setting.


\begin{table}[t]
\small
\centering
\begin{tabular}{lc}
\toprule
\textbf{} & \textbf{Normalized} \\
\textbf{Dataset} & \textbf{Orthogonal Error}\\
\midrule
FGVCAircraft & 0.008 \\
OxfordPets   & 0.007 \\
Flowers102   & 0.027 \\
StanfordCars & 0.074 \\
Food101      & 0.006 \\
DTD          & 0.055 \\
EuroSAT      & 0.024 \\
SUN397       & 0.017 \\
Caltech101   & 0.006 \\
UCF101       & 0.008 \\
ImageNet     & 0.009 \\
\midrule
\textbf{Average} & \textbf{0.022} \\
\bottomrule
\end{tabular}
\small
\caption{Orthogonality of $\mathbf{W}$ matrix: We report the normalized Frobenius norm deviation from orthogonality.}
\label{tab:orthogonality_analysis}

\end{table}

\textbf{Ablation Study}: We conduct ablation experiments to isolate the contributions of two primary design choices: Identity Initialization and the Upper Triangular constraint of the $W$ matrix. We evaluate these components on three datasets—EuroSAT (remote sensing), DTD (texture), and FGVCAircraft (fine-grained)—using a 16-shot setting.

As shown in Table~\ref{tab:ablation_study}, starting from a random initialization with an unconstrained dense matrix serves as a baseline. They struggle to achieve optimal alignment of the image-text features. While restricting the matrix to a structured upper triangular form with random weights yields a +3.24\% improvement on EuroSAT. Conversely, using identity initialization with a dense matrix ensures that the model starts at the canonical zero-shot state, but the lack of structural constraint allows for excessive manifold drift during fine-tuning.

Our proposed configuration (Identity + Upper Triangle) achieves the highest performance across the three datasets. This suggests that initializing $\mathbf{W}$ with identity preserves the pre-trained knowledge, and the upper triangular structure acts as a geometric regularization constraint. 

\begin{table}[th]
\centering
\small
\begin{tabular}{llccc}
\toprule
\textbf{Initialization} & \textbf{Structure} & \textbf{EuroSAT} & \textbf{DTD} & \textbf{Aircraft} \\
\midrule
Random & Dense & 81.04 & 69.63 & 44.88 \\
Random & Upper Tri & 84.28 & 69.57 & 44.55 \\
Identity & Dense & 82.54 & 70.74 & 45.18 \\
\textbf{Identity (Ours)} & \textbf{Upper Tri} & \textbf{85.13} & \textbf{71.01} & \textbf{45.21} \\
\bottomrule
\end{tabular}
\caption{Ablation study on EuroSAT, DTD, and FGVCAircraft (16-shot). We evaluate the impact of Identity Initialization and the Upper Triangular (Upper Tri) structural constraint.}
\label{tab:ablation_study}
\end{table} 

\textbf{Inference Time}: Inference latency due to the structured geometric transformation is negligible. Experiments show that BiCLIP requires an average of 7.78s to process 1,000 images compared to 7.77 seconds for the CLIP baseline.
\section{Conclusion}

In this work, we introduced a novel geometric perspective on adapting vision-language models through a structured bilinear transformation. By constraining the adaptation layer to an upper triangular form and initializing it with the identity matrix, we have demonstrated that it is possible to achieve state-of-the-art few-shot performance while preserving the rich, pre-trained semantic integrity of the latent space. Our findings suggest that the challenge of domain-specific adaptation is not merely one of feature extraction, but of alignment.

Through extensive angular and orthogonality analysis, we have provided a deeper understanding of the underlying structure of contrastive VLMs. We show that these models reside in a delicate "canonical" state where image and text modalities are related by latent geometric transformations. BiCLIP effectively leverages this relationship for specialized domains—such as remote sensing and fine-grained textures—by performing a controlled, non-destructive reshaping of the feature space. Finally, this research highlights that the modality gap is not a barrier for downstream tasks, but a geometric property to be navigated. By moving away from "black-box" MLP adapters toward structured, geometrically-informed heads, we can build adaptation strategies that are more parameter-efficient, mathematically interpretable, and robust to the challenges of low-data regimes. 

\newpage

{
    \small
    \bibliographystyle{ieeenat_fullname}
    \bibliography{main}
}


\end{document}